# Fire Resistance Deformable Soft Gripper Based on Wire Jamming Mechanism

Kenjiro TADAKUMA, Toshiaki FUJIMOTO, Masahiro WATANABE, Tori SHIMIZU,

Eri TAKANE, Masashi KONYO and Satoshi TADOKORO, Tohoku University, *Member, IEEE*

*Abstract—* Able to grasp objects of any shape and size, universal grippers using variable stiffness phenomenon such as granular jamming have been developed for disaster robotics application. However, as their contact interface is mainly composed of unrigid and burnable silicone rubber, conventional soft grippers are not applicable to objects with sharp sections such as broken valves and glass fragments, especially on fire. In this research, the authors proposed a new method of variable stiffness mechanism using a string of beads that can be composed of cut-resistant and incombustible metals, arrange the mechanism to form a torus gripper, and conducted experiments to show its effectiveness.

## I. INTRODUCTION

### A. Bag-Type Deformable Gripper

A robot that can perform dangerous work by remote control is required for the purpose of disaster prevention and safety of recovery workers. In particular, the gripper, which is the main end effector, is frequently used in operations that come into direct contact with the on-site environment such as valve opening and closing, switchboard operation, debris gripping and transporting as shown in Fig. 2, and each function and performance are used to perform each operation. It is an important factor that determines whether it can be implemented remotely. Unlike industrial robots, disaster-responding robots with severe power, size, and weight constraints have poor payload margins for sequentially exchanging hands according to the object to be grasped. Therefore, our research team has been researching and developing a membrane bag gripper "Omni-Gripper" that can grasp objects of any shape and size [4]. As shown on the left side of Fig. 3, the gripper is pressed against the object in a flexible state to adjust to its shape, and it is changed to a hard state by extracting the air between the powders (granular jamming transition phenomenon) to maintain the wrapping. This is a kind of soft gripper that realizes gripping. The conventional methods [1] to [3], as the powder is enclosed in the entire membrane, have a problem that the internal pressure is high and the target is pushed out. This is an epoch-making thing that realized low pressing force and large deformation.

Gripper that uses variable rigidity / flexibility switching function to change the rigidity of such structure arbitrarily has the advantage that it can grip complex shapes and fragile

*Research supported by ABC Foundation.

Kenjiro TADAKUMA, Toshiaki FUJIMOTO, Masahiro WATANABE, Tori SHIMIZU, Eri TAKANE, Masashi KONYO and Satoshi TADOKORO are with the Graduate school of Information Sciences, Tohoku University, Japan (email: tadakuma@rm.is.tohoku.ac.jp).

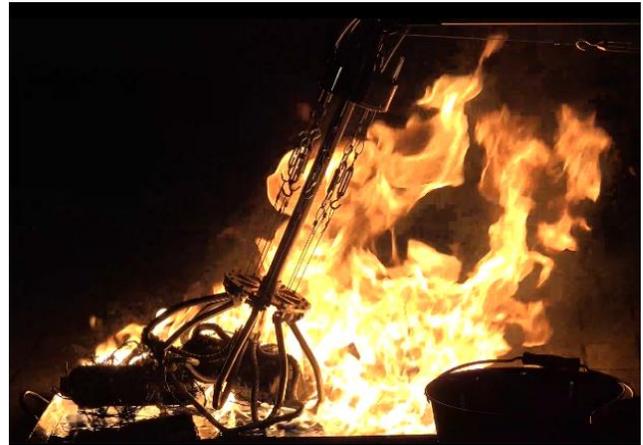

**Fig. 1: Fire-Resistant Deformable Soft Torus Gripper**

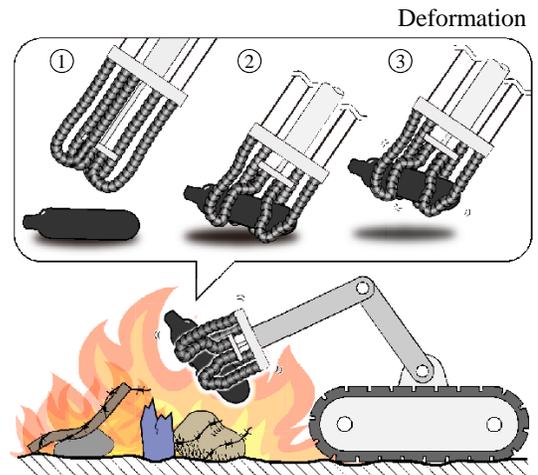

**Fig. 2: Basic concept of Fire-Resistant Torus Gripper**

objects without damaging them. This is useful for parts assembly and goods transportation in the field.

### B. Cut-Resistance Function

On the other hand, as shown in the right of Fig. 3, the conventional bag-type gripper may contaminate the environment with the powder when it flows out of the gripper when the flexible membrane is torn by contact with the sharp part of the object to be grasped. There is also a problem of loss of function without generating negative pressure due to the loss of airtightness of the space. For these reason, there has been a serious disadvantage that it was not possible to select damaged bulbs, broken glass in rubble, reinforced concrete with exposed reinforcing bars, etc. as gripping objects at the time of disaster. Therefore, our research team, as shown in Fig. 4, by protecting the rubber film with a stretchable blade-proof fabric, has realized a bag-type gripper with high cut resistance while maintaining flexibility and verified its effectiveness.

*C. Fire-Resistance Function*

Even with a bag-type hand that has been made resistant to cuts, it is extremely difficult to seize the burning debris in an ultra-high temperature environment such as a fire site after a plant explosion. This is because the flexible membrane and blade-proof fabric, which are the main elements of the mechanism, are deformed and burned in a high-temperature environment, and it is considered difficult to replace them with metals with high fire resistance. Therefore, in this study, materials were selected based on the principle of a cut-resistant gripper [17] consisting of a bead-like one-dimensional soft-rigid switching mechanism, a variable rigidity mechanism newly devised by the research team. A gripper that is compatible with modernization is constructed. Then, the effectiveness is verified through the realization of the principle proof machine shown in Fig. 1 and the actual machine experiment.

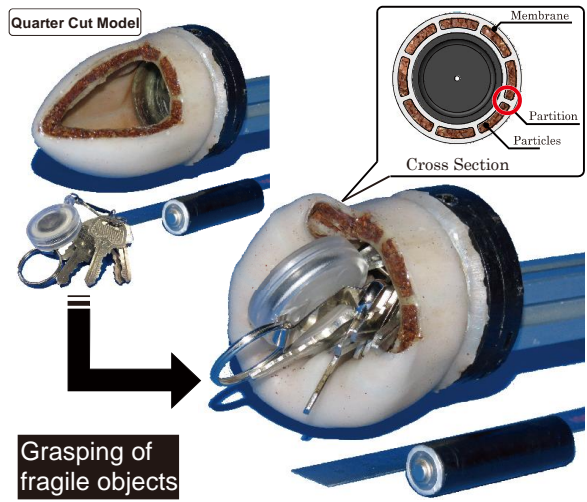

(a) Jamming membrane gripper and the inner configuration with separated wall between the space for the powder.

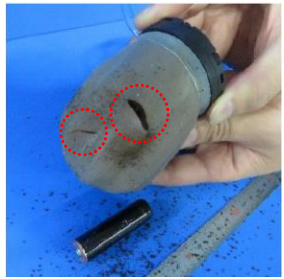

(b) Hard to grasp sharpen and pointed objects
**Fig. 3: Jamming Membrane Gripper and its Disfunction Due to Tear**

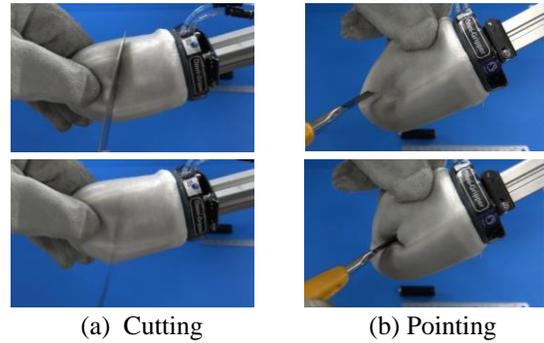

(a) Cutting      (b) Pointing
**Fig. 4: Cut-resistant Deformable Jamming Membrane Soft Gripper**

## II. BASIC PRINCIPLE OF STRING JAMMING MECHANISM

*A. Problems of the Previous Stiffness-changing Mechanism*

The conventional bag-type gripper mechanism using the granular jamming transition phenomenon as shown in Fig. 5 (a), in which the powder is enclosed in the closed space of the bag-like membrane, and the medium fluid is removed from it, so that the powders contact with each other and increases rigidity [1] ~ [5]. However, as shown in Fig. 6, there was a problem that when it was installed in a bag with an elongated ratio, buckling was likely to occur at the root even in the high rigidity mode [5]-[8]. In addition, as shown in Fig. 5 (b), a structure has been proposed in which layers of planar structures are stacked to generate a stiffness change similar to the layer jamming transition phenomenon. On the other hand, there is a problem that the radius of curvature that can be bent in the low-rigidity mode is small, and the movable range is very limited [10] [11].

In order to solve the problem of the jamming transition phenomenon with the points (powder) and surfaces (plates) as shown in Fig. 5, our research team has developed a one-dimensional flexible switching mechanism with lines (wires) as elements. This structure is composed of multiple units with beaded through holes and a central line through those holes [14]. By applying tension to the center line, the beads contact with each other with high frictional force and improve the rigidity. In the simple spherical rosary shown in Fig. 7 (a), the restoring force that minimizes the path length is generated and the posture is impaired. Therefore, the path length of the center line is independent of the posture of the mechanism as shown in Fig. 7 (b). An uneven structure that did not change was given. This structure, which has been studied as a posture fixing device for surgery [15] and a retractable arm [16], is used for stiffness switching. The actual prototype shown in Fig. 8 consists of 3D printed resin and stainless-steel wire. The tensile force applied on the wire depending on the strength of the material. Since the structure does not have a flexible membrane, it has achieved a high level of cut resistance at the principle level in that it does not inherently cause the phenomenon of tearing by cutting.

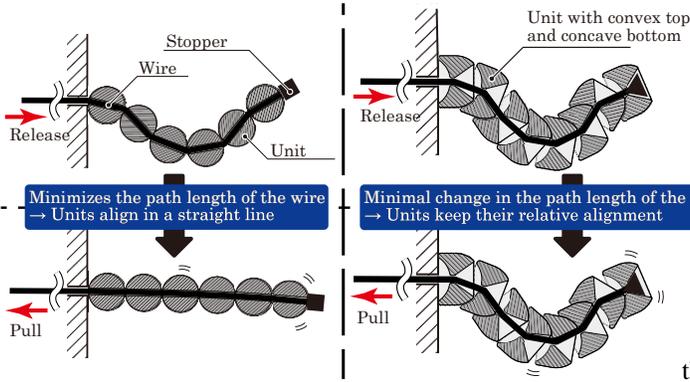

**Fig. 5: Principle diagram of proposed
One-Dimensional Variable Stiffness Mechanism**

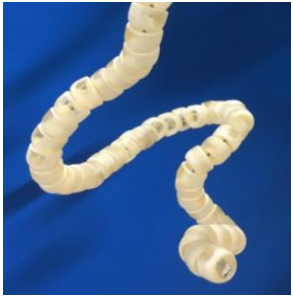

**Fig. 6: Proof-of-principle model of
One-Dimensional Variable Stiffness Mechanism**

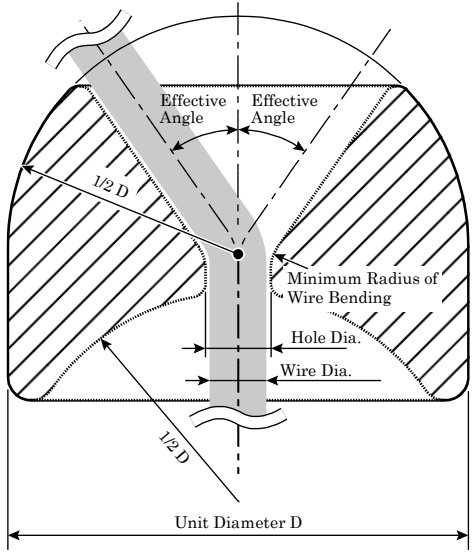

**Fig. 7: Single Cup-shaped Bead Model for
One-Dimensional Variable Stiffness Mechanism**

Fig. 7, 8 shows the cross section of the unit. If the diameter is specified in a shape like a ball joint, the curvature of the curved surface of the ball joint can be determined. The shape can be determined by five parameters: diameter, effective angle of the orientation, wire diameter, hole diameter, and wire bending radius. The radius and diameter of the bead are set as $R_1$, $D_1$, respectively. The diameter at the former convex part is set as $R_2$ and that of rear part of the concave part is set as $R_3$. There are relationships between these diameter as following equation (1).

$$R_2 = R_3 = R_1 \ (=D_1/2) \quad ------(1)$$

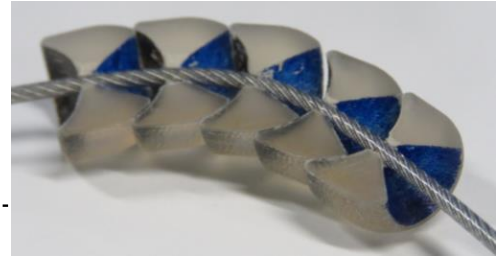

**Fig. 8: Real Cut Model of the Single Cup-shaped Bead**

The minimum radius of the wire bending is show as $r_2$, the curvature radius at the inner surface of the hole of the beads set as $r_1$. Then there should be the relationship based on the equation(2). The diameter of the rear hole of the beads is set as $SD_1$ and that of the wire is set as $SD_2$, and small clearance for smooth motion is set as e, so there should be the relationship as following equation (3).

$$r_1 > r_2 \quad ------(2)$$
$$SD_1 > SD_2 + e \quad ------(3)$$

In the one-plane constrained method described later, it is possible to limit the bending area by providing the movable range only in one direction.

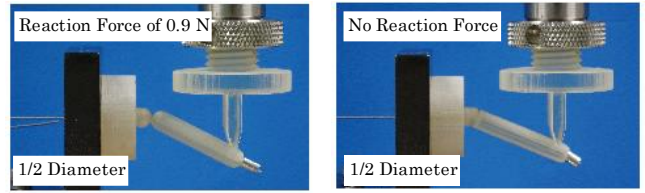

(a) Simple Sphere Structure    (b) Convex and Convex Structure
**Fig. 9: Comparison between Normal Ball Shaped Type
Bead and Cup Shaped Type one**

Using a half-size unit with a diameter of 6 mm, measurements were taken at only one of the basic joints as shown in Fig.9. The deformation of about 30 ° was given, and then the force to return to a straight line was measured by pulling the wire. As experimental conditions, the cylinder pressure was 100 kPa and the wire tension was 41 N, and 10 measurements were performed. As a result, it was confirmed that the reaction force was not generated in 0.9 N for the simple sphere structure and the uneven structure.

## III. DEVELOPMENT OF THE MECHANICAL PROTOTYPE MODEL OF THE DEFORMABLE SOFT GRIPPER WITH WIRE JAMMING MECHANISM

*A. Deformable soft gripper with torus configuration based on string Jamming Mechanism*

Our research team has devised a cut-resistant gripper that can grasp sharp objects by constructing a torus shape by discretely arranging a one-dimensional soft-rigid switching mechanism in the shape of a finger [17]. In this study, we will reconstruct this and optimize the structure to obtain fire resistance. In order to limit the familiar direction to the radial direction, a one-dimensional flexible switching mechanism constraining the degree of freedom of bending to 1 was applied, and the pipe from the root to about half was composed of a rigid pipe. The flexible switching operation is realized by driving an equalizer with an air cylinder that can draw each wire with equal force regardless of the posture as shown in Fig10. In addition, when releasing the grasped object, if the torus continues to be depressed, it will interfere with the grasping of the next object. So it was connected to the tip fastening part of the switching mechanism to give a restoring force. Furthermore, a free rotating shaft was provided at the base of the one-dimensional flexible switching mechanism and a torsion spring was added to enable opening and closing in the radial direction.

As a result, when the air cylinder is pressurized in the opposite direction to restoration, a gripping operation and an entraining operation are generated by opening and closing, and a force that actively adapts to the object is generated. It is now possible to hold objects The flexible switching unit is attached to the tip of the hinge and the ends are fixed to the cylinder rod as a whole. There is a spring in the hinge, which always exerts a force to open it. By applying pressure to the piston B port of the equalizer, the wire is pulled and the flexible / rigid switching unit is cured as shown in Fig.11. At the same time, it moves toward the closing side of the hinge and grips with familiarity. The A port is pressurized at a constant pressure of about 10-50 kPa and has the role of a constant load spring. The cylinder is always pushed outwards to maintain the initial state of the flexible switching unit. It also plays a role of deformation to the initial state after the object is gripped and released. The overview of the first prototype model of the deformable soft gripper based on wire jamming mechanism is shown in Fig.12.

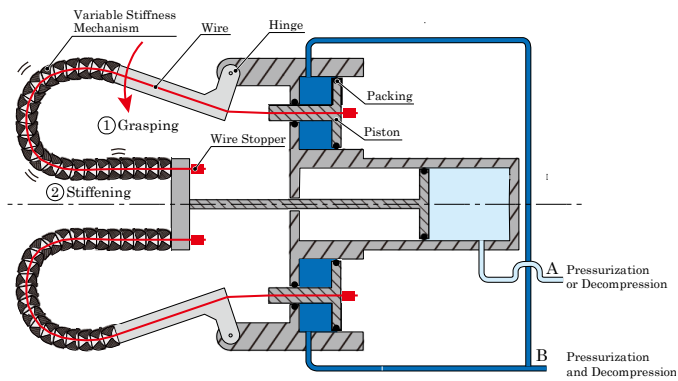

**Fig. 10: Basic Configuration of the Deformable Soft Gripper Based on Wire Jamming Mechanism**

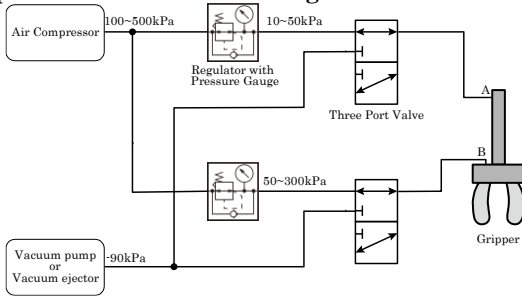

**Fig. 11: Pneumatic Circuit Model for One-Dimensional Torus Gripper**

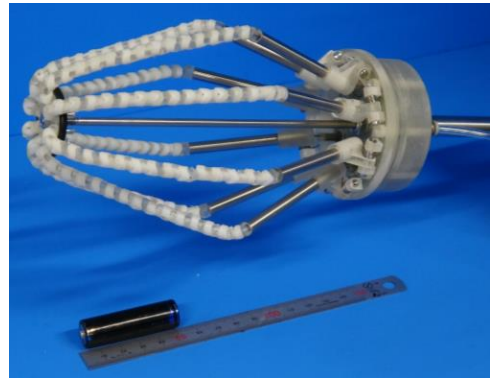

**Fig. 12: Proof-of-principle the deformable soft gripper based on wire jamming mechanism**

**Table1: Specification of the First Prototype Model of the Wire Jamming Gripper**

| | | |
|---|---|---|
| 1-line | Number of units | 30 |
| | Diameter | 6 [mm] |
| Whole | Number of lines | 8 |
| | Diameter | 120 [mm] |
| | Length | 350 [mm] |
| | Tensile force | 37 [N] |
| | Stroke | 140 [mm] |

The A port and B port cylinders are driven using compressed air and vacuum. It is better to operate with compressed air from both cylinders, such as a return cylinder, but the number of airtight parts increases, the number of tubes becomes two, and multiple pipes are required. Therefore, it is configured like a single-acting cylinder, vacuum pressure is used instead of the return spring, and when the vacuum pressure is applied, the cylinder moves in the opposite direction when pressurized. This simplifies the mechanism. In order to verify the basic gripping performance of the devised gripper, a preliminary experiment was conducted using a small prototype. Specifically, it handled bottles with large gripping parts, objects with flat parts such as tapes and boxes, objects with sharp parts such as scissors, and operations such as valves and vice. Fig.13 shows how the sharp part of the scissors and knife is gripped on each. By grasping after being pressed against the object, it fits into the object shape. It was possible to grip the blade edge directly during gripping, and the effectiveness of the sharp object was demonstrated.

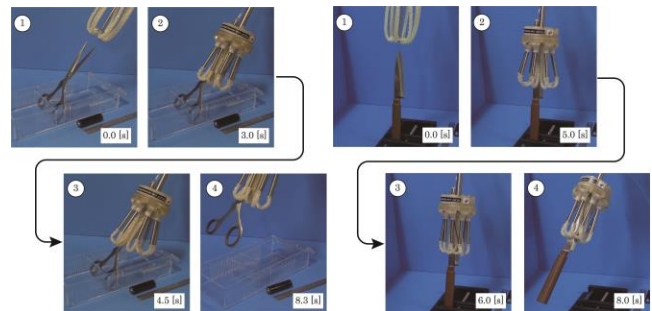

(a) Scissors   (b) Knife

**Fig. 13: Bottom View during the Grasping Two Types of Objects**

## IV. BASIC GRASPING PERFORMANCE EXPERIMENTS

### A. Experimental Setup

A basic gripping experiment was conducted as shown in Fig. 14-16, in which a triangular prism or cylinder as a sharp object is gripped as a test object and observed with the camera from the side and bottom, and basic knowledge is obtained. The displacement and force after gripping are measured with a tensile tester. Figure 15 shows the cylinder and triangular prism used in the experiment.

i) Experimental conditions:
  Air pressure: 100 kPa, 200 kPa,
  Gripping object (cylinder) diameter 30 mm, material: polyacetal,
  Grip object (triangle) tip angle 30 °, material: acrylic

ii) Experimental Procedure:

The gripper is lifted after gripping a fixed gripping object, and the maximum gripping force and gripping characteristics are measured from the displacement and force at that time. Each measurement was performed 10 times. The state of the experiment is shown in Fig. 14. The left image is from the side camera, and the right image is from the bottom camera into the result in Fig .16. After the gripper was pressed against the object to be gripped and gripping was performed, force and displacement measurements were started, the gripper was lifted at a constant speed, and the situation at that time was observed.

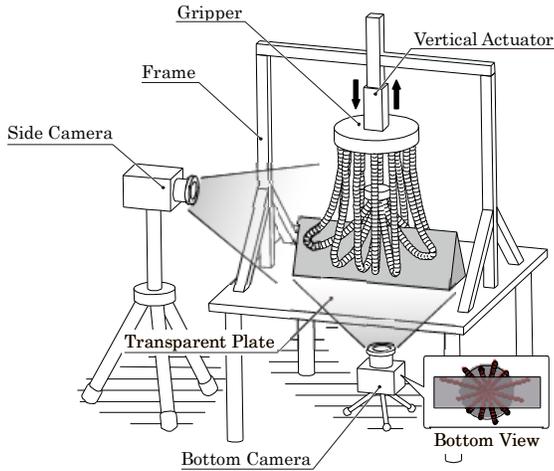

**Fig. 14: Experimental Setup for the Gripper to Grasp Sharpen Object**

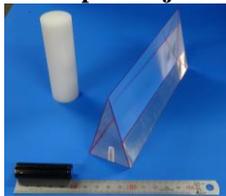

**Fig. 15: Cylinder Object Sharpen Triangle Object for Grasping Experiments**

*B: Cylindrical Object:*

A graph of the experimental results is shown in Fig. 17. The horizontal axis shows the upward displacement, and the vertical axis shows the force at that time. Both the cylinder and the triangle can be confirmed to follow the shape change due to the grip force. As a consideration, in the cylinder gripping, the improvement of the gripping force was confirmed according to the pressure. The pressing force can also be confirmed at the left end. The part that vibrates at 200 kPa is thought to be due to repeated sliding with the object. The part where the gripping force in the second half is negative is thought to be because the gripping object is pushed out by the closing force of the hinge.

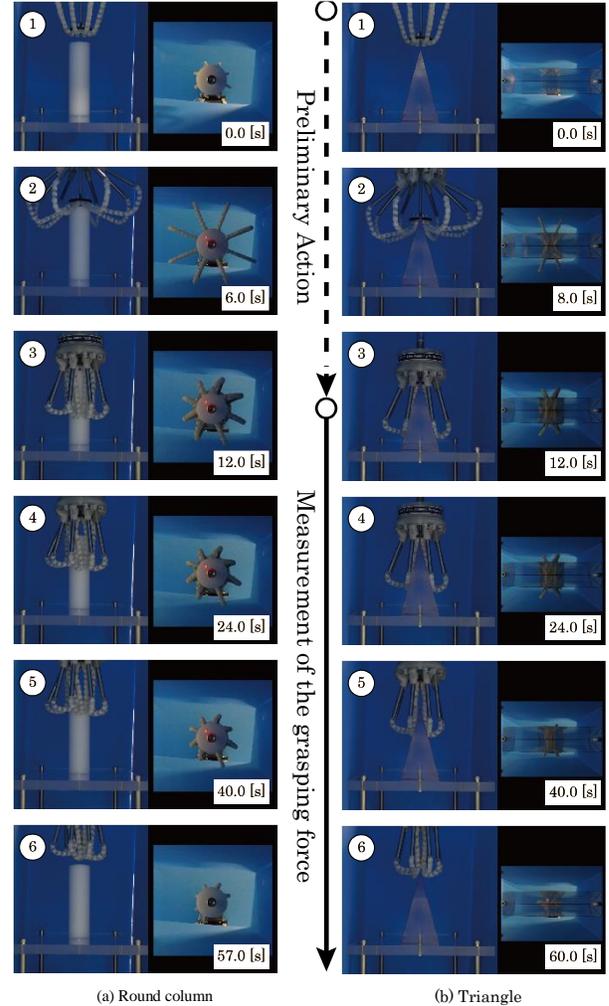

**Fig. 16: Bottom View during the Grasping Two Types of Objects**

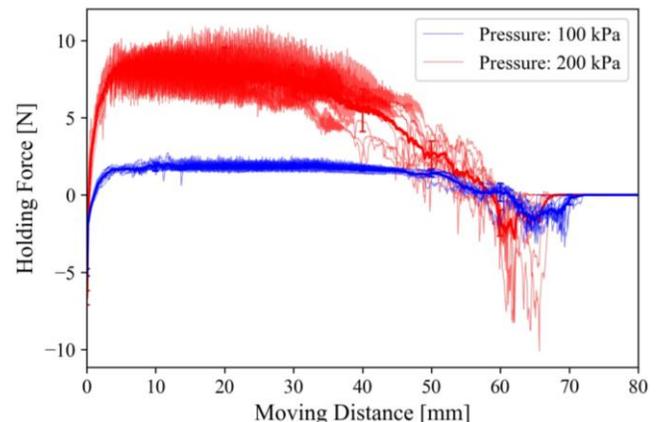

**Fig. 17: Experimental Result of the Holding Force of the Cylindrical Object Gripping**

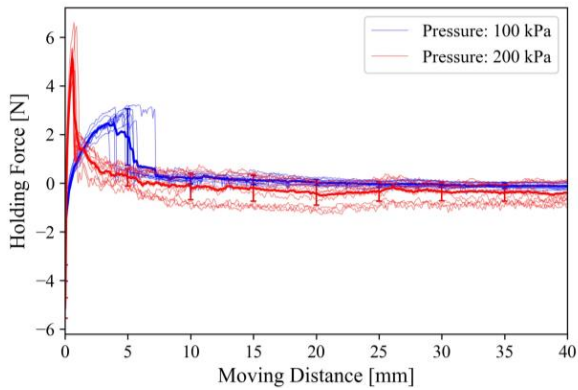

**Fig. 18: Experimental results of triangle Holding force**

*C: Triangle as a Sharpen Object:*

The result of the triangular gripping force experiment is shown in Fig. 18. This is similar shape of that of the cylindrical one, the gripping force was improved at 200 kPa. Since the shape is triangular, it can be confirmed that the gripping force decreases as it is lifted. Since the triangle was cut out of acrylic, the acute angle part was sharp, but the gripper could be gripped without breaking.

*D: Pressing Force*

Next, we measured the force when pressing the gripper against the object. The smaller this force is, the easier it is to grip a fragile object or an airborne object that is difficult to exert.

Experimental conditions: press on the cylindrical shape, and change pressure to A port (10,20,50 kPa). The displacement 0 is determined based on the maximum gripper length. And the displacement is determined by gripper stroke (140 mm).

Experimental procedure: at first the start measurement with the gripper away from the cylinder, lower the gripper, and press it against the object to be gripped. Lift it down and move to the initial position. This operation was repeated 10 times each.

Experimental result: a graph of the experimental results is shown in Fig. 19. The horizontal axis shows the displacement of pressing, and the vertical axis shows the force at that time. The pressing force increases as the pressure increases. In each graph, the rising position is the gripper contact position with the gripping object. The lower the pressure is, the more the point shifts. At low pressure, the extension force of the cylinder loses the spring force of the gripper hinge. This is because the tip of the gripper does not extend completely. At this time, the stroke of the gripper is reduced. The gentle rise from the contact at 50 kPa to about 60 mm is thought to be due to this spring. The upward trend at the right end in all results is the effect of the force acting laterally on the cylinder rod due to the shortened A port cylinder.

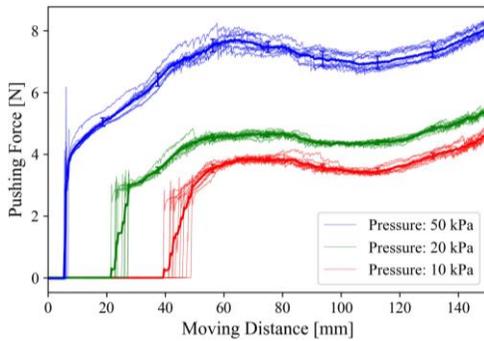

**Fig. 19: Experimental Result of the Relationship between Moving Distance and Pushing Force**

Fig. 20 shows the experiment using the membrane gripper.
i) Experimental conditions:
  Membrane gripper Measurement was performed with the gripping object: triangle, vacuum pressure: -90 kPa, push-in amount: 40 mm, speed: 100 mm / min, and the point at which the tip touched the gripping object in the natural state of the gripper as a displacement of 40 mm. Number of experiments: 10
ii) Experimental procedure:
  The gripper was pressed against the object to be gripped by the push-in operation, and the vacuum state was switched when the push-in amount was reduced. After that, measurement was started and the lifting operation was performed. After rising to the initial displacement, the pressure was switched to atmospheric pressure.
iii) Experimental results and discussion:
  The experimental results are shown in the graph in Fig. 21 The blue color is a plot of the experimental results of the conventional film jamming gripper, and the red color is the result of superimposing the results of the torus gripper utilizing the flexible switching mechanism. Compared to the conventional one, it shows a strong gripping force about 1.4 times.

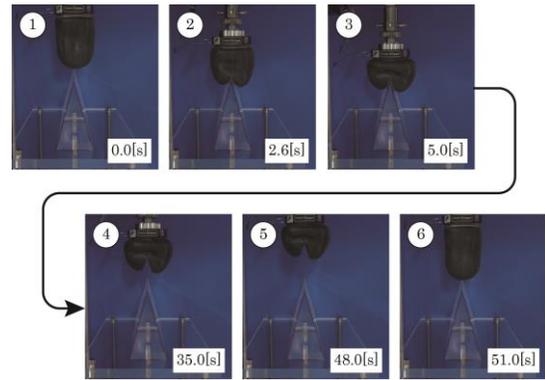

**Fig. 20: Bottom View during the Grasping Two Types of Objects**

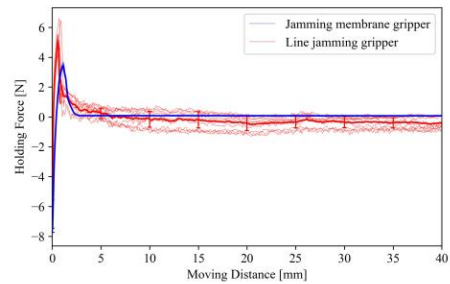

**Fig. 21: Comparison of experiment results of gripping force**

*E. Comparison of pressing force with membrane*
i) Experimental conditions:
  Membrane gripper measurements were taken with the object: triangle, push-in amount: 40 mm, speed: 100 mm / min, and the point at which the tip touched the gripping object in the natural state of the gripper as 10 mm displacement. Number of experiments: 10
ii) Experimental Procedure:
  The gripper was pressed against the object to be gripped by the push-in operation, and the vacuum state was switched when the push-in amount was reduced. After that, measurement was started and the lifting operation was performed. After rising to the initial displacement, the pressure was switched to atmospheric pressure.

iii) Experimental results and discussion:

Fig. 22 shows the pressing force of the film jamming gripper against the cylinder. The blue plot is the result of the conventional film jamming gripper, and the red plot is the result of the linear torus gripper. In both grippers, the origin was set 10 mm above the object, so contact started at a displacement of 10 mm in the graph and a pressing force was generated. With linear torus grippers, the increase rate of pressing force is low, and when it is 25 mm or more, the pressing force is lower than that of membrane jamming grippers. Therefore, the linear torus gripper is considered more advantageous for gripping that requires a large amount of torus folding.

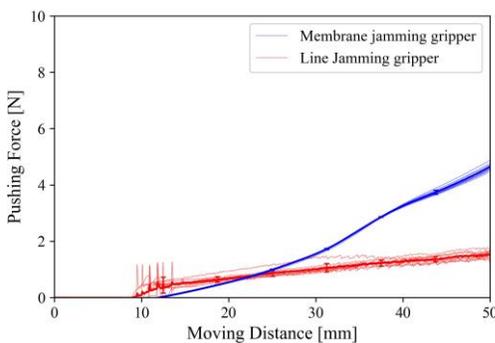

**Fig. 22: Comparison of experimental results of pushing force**

In this section, a basic torus gripper experiment was conducted. In the object grasping experiment, an object with sharp parts such as scissors and knives was grasped. In the experiment, the gripping force characteristics and the pressing force of the basic shape were confirmed. In the experiment, the gripping force characteristics were measured using a basic shape cylinder and triangle. A comparative experiment with a conventional jamming film gripper was conducted, and the effectiveness of this mechanism was confirmed.

## V. FIRE-RESISTANCE FUNCTION

### A. Realization of Fire-Resistance Function on the Gripper

The principle proof machine of the fire-resistant torus gripper mechanism which has been made fire-resistant and enlarged by following the structure of the gripper shown in Fig. 23 with the drawing and real protype model. If there is no fire-resistance function, the gripper should ignite and spread out as shown in Fig .24. The main structural materials consisted of aluminum alloy and stainless steel. The rosary was made of fire-resistant, high-strength, lightweight titanium alloy, and the center line was made of tungsten wire with excellent heat resistance and high tensile strength. Since the O-ring used to seal the equalizer remains made of rubber with low heat resistance as shown in Fig. 25.

An equalizer and the hand are made independent to isolate it from the gripping object, which is a heat source, and the wires extending from each are turned. It is configured to connect with a buckle. In addition, a cavity was provided in the holder to allow cooling water to be circulated so that the equalizer could be cooled as needed.

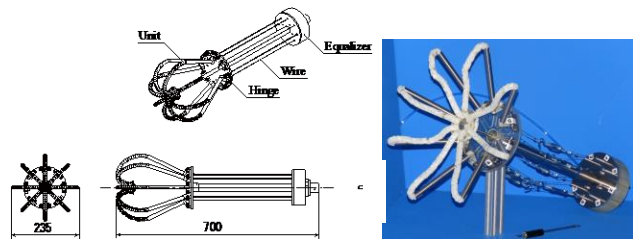

**Fig. 23: Design diagram and Open-finger state of Fire-Resistant Torus Gripper**

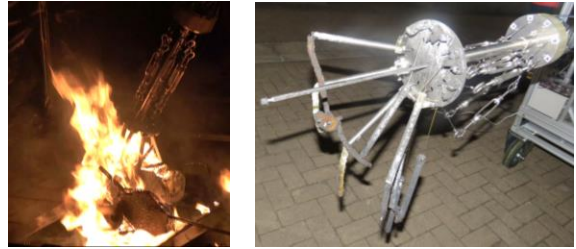

**Fig. 24: Gripper without any fire-resistance should ignite and spread out**

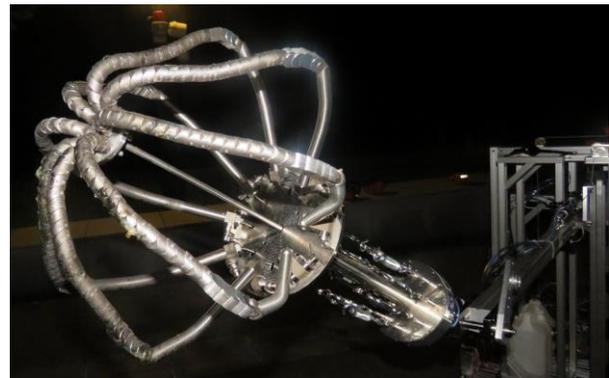

**Fig. 25: Proof-of-principle model of Fire-Resistant Torus Gripper**

### B. Basic Experiment

A gripping experiment was carried out by placing the embodied proof-of-concept machine on a handcart equipped with a windshield for protecting the operator, and this was operated by a human to wind the barbed wire and supply fuel as shown in Fig. 26. Fig. 27-28 shows how about 2 [kg] of concrete rubble that was applied and burned up was caught in a fire and transported to an external fire extinguishing facility (bucket). As can be seen in Fig. 27-28, which shows a magnified image of the hand, the effectiveness of a cut-resistant and fire-resistant gripper mechanism that can handle an object without damaging the sharp part or losing its function due to a burning environment has been confirmed. did it. It was also shown that the characteristics of a soft gripper mechanism that does not deform or destroy the object to be gripped are maintained. Details of the self-weight / torque compensation mechanism for the bogie and arm joint will be reported in a later report.

## VI. FOR MORE PRACTICAL APPLICATION

By using the refractory gripper demonstrated in this study, it becomes possible to grip high-temperature, sharp, indeterminate, fragile, and weak objects that were previously impossible. This is because the water gun is installed in the center of the robot hand, and the debris, doors, and lids are opened, and the fire extinguishing agent is sprayed directly on the fire source ahead to apply it to new fire-Fighting equipment that extinguishes quickly and reliably. Means that is also possible. Generally, in order to improve cut resistance and fire resistance, the structural material must be made of metal as in this study, and the weight of the gripper becomes a problem. Especially when mounted on firefighting robots and disaster response robots, not only compensates for the power source weight and torque, but also reduces the weight of the mechanism itself, so as not to cause the debris to be stepped on or to lose its posture due to unstable scaffolding. Is also essential.

## VII. CONCLUSION

In this study, we studied the damage on a soft material due to complex, sharp, and fragile shape of grasping object and high-temperature in a burning environment was extremely difficult for a conventional soft gripper using a flexible switching mechanism. Therefore, we invented a one-dimensional flexible switching mechanism consisted of a rosary and a center line. A fire-resistant torus gripper mechanism that enables gripping in a restrained state was devised, and the effectiveness of the devised principle was demonstrated through the realization of the proof-of-principle machine and the gripping motion experiment.

In the future, not only in the field of fire, but also in the industrial field that handles high-temperature objects such as the steel industry, and in the field of applications such as inspection and decommissioning work in nuclear power plants where it is difficult to use rubber materials that are easily deteriorated by radiation. We will look at it widely and refine it for practical use.

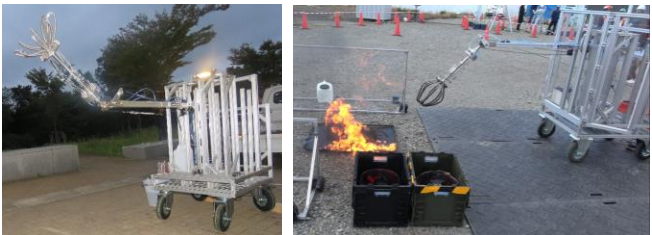

**Fig. 26: Fire-Resistant Torus Gripper on handcart with weight and torque compensation mechanism**

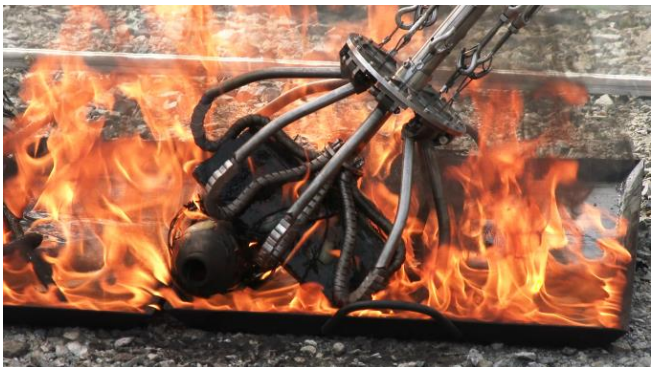

**Fig. 27: On-fire grasping test of Fire-Resistant Torus Gripper (Please see the attached file to understand how deformable this gripper is)**

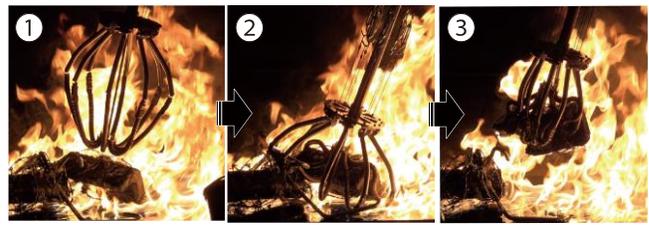

**Fig. 28: Magnified view of the on-fire grasping test**